# Tamil Vowel Recognition With Augmented MNIST-like Data Set

Author: Muthiah Annamalai

## Abstract:

We report generation of a MNIST [4] compatible data set [1] for Tamil vowels to enable building a classification DNN or other such ML/AI deep learning [2] models for Tamil OCR/Handwriting applications. We report the capability of the 60,000 grayscale, 28x28 pixel dataset to build a 92% accuracy (training) and 82% cross-validation 4-layer CNN, with 100,000+ parameters, in TensorFlow. We also report a top-1 classification accuracy of 70% and top-2 classification accuracy of 92% on handwritten vowels showing, for the same network.

## 1. Introduction

Indian languages have complex orthography; particularly the Tamil script uses a circular orthography structure in stark contrast to the Roman/Latin alphabets which may be angular and upright. OCR work in Tamil previously has used various classic pattern-recognition techniques like matched filtering, correlations, etc. [8,9] and used transfer-learning deep-learning techniques to this area, based on handwritten dataset OCR [3]. Similar work, on MNIST compatible [4], handwritten Kannada (Indian language) numerals identification was presented earlier [5]. Since each of these approaches require creation of the dataset by painstaking data normalization and data-collection from varied sources our approach presents an automatic-bootstrapping of datasets for Tamil OCR using existing fonts and a data-augmentation algorithm.

We present 60,000 28x28 pixel images of 12 vowels of the Tamil alphabet [அ,ஆ,இ,ஈ,உ,ஊ,எ,ஏ, ஐ,ஒ,ஓ,ஔ] and the *aytham* letter ஃ, as a data source for 13-multiclass classification. We perform the training for this dataset using TensorFlow [6] and validate the performance. This system, methodology and results are discussed below.

## 2. Dataset Preparation

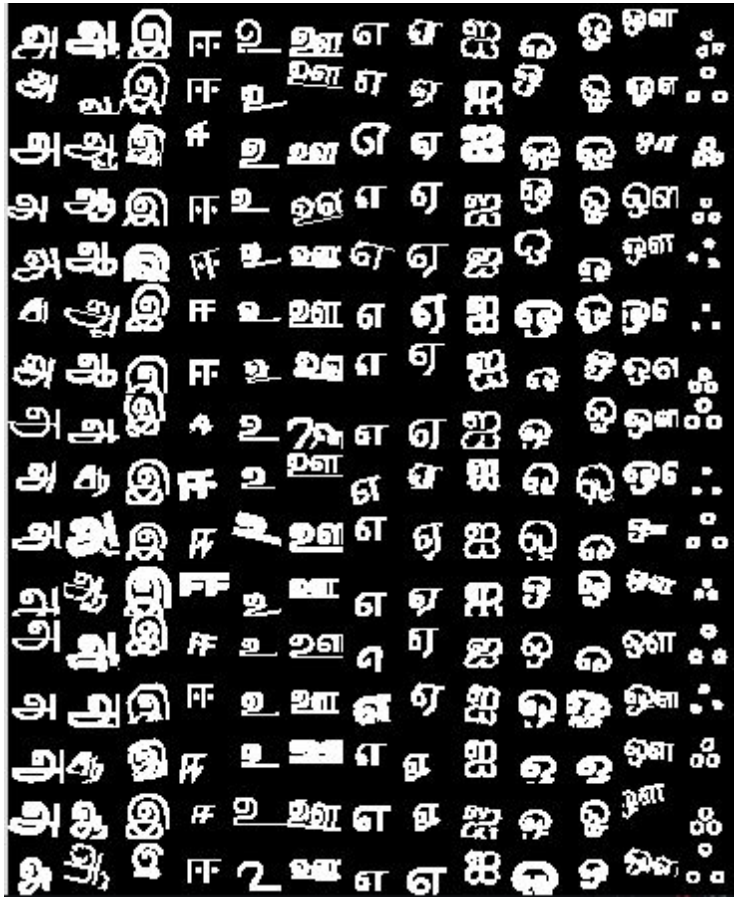

Fig. 1: Sample of vowels from the dataset. Some images shown

### 2.1 Fonts

The following 35 fonts were chosen for their public availability or open-source license on many systems to generate the augmented dataset. These contain popular fonts like InaiMathi (by Murasu Anjal team shipped in Apple Mac OS-X), Arial, Latha (by Microsoft), as well as open-source fonts like Catamaran etc. Many of the fonts in our list are picked from the Thamizha Open-Source font-database [7].

| | |
|---|---|
| 1. InaiMathi-MN | 20. 1094_ChemmozhiComic |
| 2. Kavivanar-Regular | 21. 1094_ChemmozhiParanar |
| 3. Lohit-Tamil | 22. 1094_ChemmozhiThendral |
| 4. MeeraInimai-Regular | 23. 1094_ChemmozhiThenee |
| 5. MeeraTamil-Regular | 24. 1094_ChemmozhiTimes |
| 6. MuktaMalar-Regular | 25. 1094_ChemmozhiVaigai |
| 7. Pavanam-Regular | 26. Akshar |
| 8. PostNoBillsJaffna-Regular | 27. Arial Unicode |
| 9.  RRJanaTamil<br>10. Tamil MN<br>11. Tamil Sangam MN<br>12. ThendralUni<br>13. TheneeUni<br>14. Uni-Tamil042<br>15. Uni-Tamil046<br>16. Uni-Tamil150<br>17. Uni-Tamil195<br>18. VaigaiUni<br>19. YaldeviJaffna-Regular | 28. ArimaMadurai-Regular<br>29. BalooThambi-Regular<br>30. Catamaran-Regular<br>31. Coiny-Regular<br>32. HindMadurai-Regular<br>33. aava1<br>34. Latha<br>35. tau1_bar |

## 2.2 Transformations

A 35% of the images generated in the augmented dataset are rotated and another 15% of images are rotated and translated. Rotation range is ±15° and translation by ±5 pixels applied to the dataset are further clipped to the 28 px square area. These series of transformations, with bilinear interpolation, provide a sufficiently true but noisy dataset for building a true model of vowels.

## 2.3. Outliers

The transformations and varied nature of the fonts kerning, widths, serif, and unique aspects of *vattezhutu* (வட்டெழுத்து) - Tamil orthography - certain vowels spill out of the chosen square area. Particularly vowels ஔ, and the *aytham* letter ஃ were seen to spill out of the square area more frequently and hence assigned as smaller rendering font size than the corresponding vowels. Suitable considerations are taken during training and cross-validation to eliminate overflow images from being model inputs. Upon cleanup we find 48,465 training samples as meaningful in one set of 60,000 images and 52,481 training samples are meaningful in the next set of 60,000 images.

# 3. Classification Model

## 3.1 Software Setup

The software setup used TensorFlow 2.0, Numpy, Python Imaging Library PIL, all on Python 3.6 running on Mac OS-X High Sierra, in a 4GB RAM core i5 processor. While this setup was not ideal to run accelerated compute, this yielded converging results all the same. Training was run using the built-in Adam optimizer on a 75-25 split training and test data based on [1], with batch size of 128 images and conservative learning rate.

**Table: 1**. TensorFlow Lite version of Fully Connected and CNN models built from our dataset.

| Fully-connected model (TensorFlow Lite) | CNN model (TensorFlow Lite) |
|---|---|
| Parameters: 1,335,309 | Parameters: 116,109 |
| 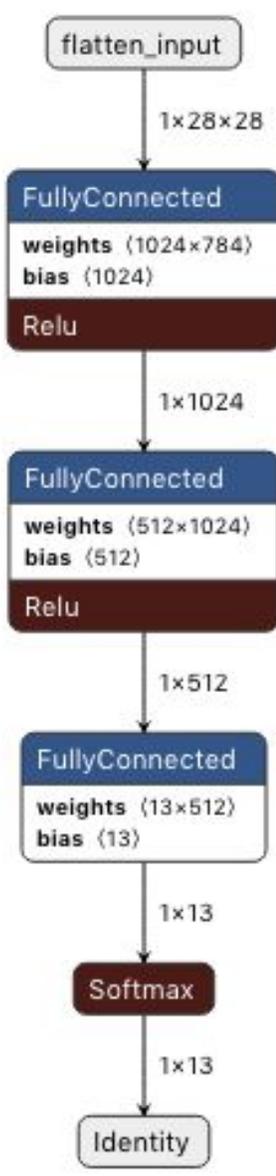 | 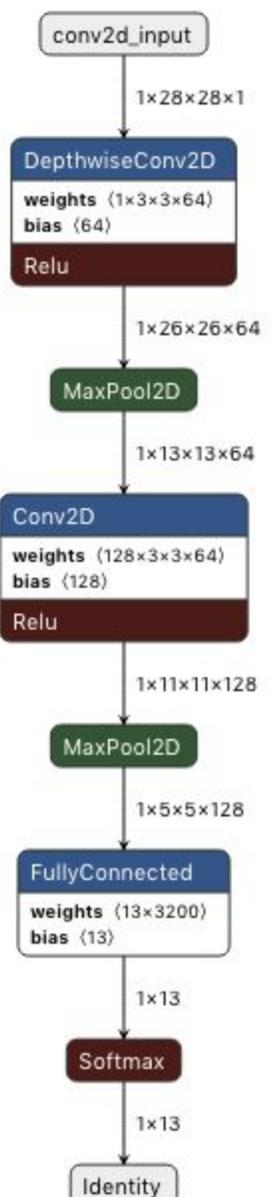 |

## 3.2 Fully Connected Neural Network

We used three fully-connected "dense" layers to classify the image into the correct label among 13 classes. This yielded a 89% accuracy on training dataset and 80% accuracy on cross-validate dataset, with performance plateauing on loss across number of epochs cross 500. This is somewhat in good company of wild success of CNN [2] in today's landscape that representational learning via dense/fully-connected networks is much more limited than their convolutional counterparts.

Our trained network parameters are shown in Table. 2, and the TensorFlow-Lite architecture is shown in Table. 1 (left). This network is also about 10x larger than the corresponding CNN shown in Sec. 3.3.

**Table: 2**. Fully Connected Model layers with 1,335,309 total trainable parameters.

| Layer (type) | Output Shape | Param # |
| --- | --- | --- |
| flatten (Flatten) | (None, 784) | 0 |
| dense (Dense) | (None, 1024) | 803840 |
| dense_1 (Dense) | (None, 512) | 524800 |
| flatten_1 (Flatten) | (None, 512) | 0 |
| dense_2 (Dense) | (None, 13) | 6669 |

## 3.3 Convolution Neural Network

We use 2 CNN layers followed by a flatten and fully-connected + softmax layer for this network architecture to report our best yet performance on this dataset.

The initial 2 CNN layers had fewer feature map, 32 and 64 respectively on layers 1, and 2 yielding a net performance of training accuracy at 85% and saturating loss with increasing epoch; saturation of the training and cross-validation dataset accuracy at 65% guided us to increase more parameters in the network.

Upon increasing the feature map sizes to 64 and 128 respectively and re-running the training we could see the model converge to a higher plateau of accuracy and cross-validation.

**Table: 3**. CNN Model layers with 116,109 total trainable parameters.

| Layer (type) | Output Shape | Param # |
| --- | --- | --- |
| conv2d (Conv2D) | (None, 26, 26, 64) | 640 |

| max_pooling2d | (None, 13, 13, 64) | 0 |
| --- | --- | --- |
| conv2d (Conv2D) | (None, 11, 11, 128) | 73856 |
| max_pooling2d | (None, 5, 5, 128) | 0 |
| flatten (Flatten) | (None, 3200) | 0 |
| dense (Dense) | (None, 13) | 41613 |

The trained model shows accuracy of 92% in the training dataset and a cross-validation accuracy of 85%. This model is also converted to TensorFlow-Lite and visualized in Table. 3 (right).

## 4. Results

We use data from handwritten input to see if our font-based augmented data trained classification model has created a suitable representational learning [2] of the Tamil vowel shapes. To this end normalize the 128x128 pixel images to MNIST compatible grayscale images in 28x28 pixel shape and compare the inputs to predicted categories - results are shown in Table 4. We use the CNN model from section 3.3 above.

The top-1 accuracy of ~ 70%, containing 4 errors in 13 labels, is shown in the Table. 4. Specifically, ஒள misclassified as ஊ, ஓ misclassified as ஐ, எ misclassified as ஒ, and இ misclassified as ஆ. However, upon closer investigation we find the top-2 accuracy has only one error ஓ -> ஐ is mistaken more than 2 spots away and other cases line up for a total of 92%. This is a signature of the representation learning in our CNN model from an augmented dataset carrying over into successful prediction of the handwriting dataset.

Comparative performance of the Fully-Connected model, Sec 3.2, shows an accuracy of only 62% on this same handwritten dataset, despite the model being 10x larger in terms of number of trained parameters.

**Table: 4**. Top-1 accuracy of 4-layer CNN model on handwritten letter dataset at 70% accuracy. Prediction legend (color-online): Red - incorrect classification, Green - correctly classification.

| Input Image | Prediction | Input Image | Prediction |
| --- | --- | --- | --- |
| அ | அ | எ | எ |
| ஆ | ஆ | ஏ | ஒ |
| இ | ஆ | ஐ | ஐ |
| ஈ | ஈ | ஒ | ஒ |
| உ | உ | ஒ | ஐ |
| ஔ | ஔ | ஔ | ஔ |
| | | ஃ | ஃ |

We suspect a much better outlier avoidance algorithm could provide noise free dataset and avoid the cases of the misclassification and can be a task in future studies.

## 5. Conclusion

We have created the first open-dataset for Tamil vowel classification in compatibility with MNIST dataset for OCR application/handwriting model construction; this may also serve as an easy introduction to Tamil AI/ML model construction.

The successful results from our classification model studies show no limitations for extending our approach, from 13 labels, to the full alpha-syllabary set of 247 (12 + 18 + 12x18 + 1) strictly Tamil letters (and about 343 letters if we include Grantha Sanksritized letters) with bootstrapped data in quest for an on-line handwriting recognition system.